\documentclass[10pt, conference, compsocconf]{IEEEtran}
\usepackage{cite}

\ifCLASSINFOpdf
\usepackage[pdftex]{graphicx}
\else
\fi
\usepackage[cmex10]{amsmath}

\usepackage{multirow}

\usepackage{url}

\begin{document}
%
\title{A Multi-Task Network for Localization and Recognition of Text in Images}


\author{\IEEEauthorblockN{Mohammad Reza Sarshogh}
\IEEEauthorblockA{Capital One\\
McLean, Virginia\\
reza.sarshogh@capitalone.com}
\and
\IEEEauthorblockN{Keegan E. Hines}
\IEEEauthorblockA{Capital One\\
McLean, Virginia\\
keegan.hines@capitalone.com}
}


%


\maketitle

\begin{abstract}
We present an end-to-end trainable multi-task network that addresses the problem of lexicon-free text extraction from complex documents. This network simultaneously solves the problems of text localization and text recognition and text segments are identified with no post-processing, cropping, or word grouping. A convolutional backbone and Feature Pyramid Network are combined to provide a shared representation that benefits each of three model heads: text localization, classification, and text recognition. To improve recognition accuracy, we describe a dynamic pooling mechanism that retains high-resolution information across all RoIs.  For text recognition, we propose a convolutional mechanism with attention which out-performs more common recurrent architectures.   Our model is evaluated against benchmark datasets and comparable methods and achieves high performance in challenging regimes of non-traditional OCR. 
\end{abstract}


%
\IEEEpeerreviewmaketitle

\section{Introduction}
The extraction of text from images is an important challenge for many applications including document analysis, scene understanding, and automated driving, among others. In certain instances, it is desirable to extract text objects from naturalistic scenes such as those of roadways and cityscapes \cite{DBLP:journals/corr/JaderbergSZK15, DBLP:journals/corr/abs-1801-01671}. In other instances, text might be extracted from documents and images of documents. This latter case has traditionally been called Optical Character Recognition and has been a focus of many successful approaches \cite{DBLP:journals/icdar/Smith2007}.

Recent advances in computer vision have demonstrated promising success in detecting text in naturalistic scenes \cite{DBLP:journals/corr/ShiBY15, DBLP:journals/corr/LiaoSBWL16, DBLP:journals/corr/JaderbergSVZ14a, DBLP:journals/corr/abs-1801-01671}, and many of these approaches build upon previous work in object detection \cite{ DBLP:journals/corr/LiuAESR15, DBLP:journals/corr/HeGDG17}. It has been noted that this regime poses unique challenges that are not faced in traditional OCR, including background/object separation, multiple scales of text detection, text orientation, coloration and occlusion. These methods have proven useful for scene understanding and image retrieval.

Less well studied, however, is the regime of text extraction from images of complex documents. This case falls between “traditional” OCR (where well-specified documents are laid out in a clear fashion) and scene-based text detection (where a small number of relatively large text boxes are extracted from images and video). The regime of interest here is one of  complex and unstructured documents, for example images of receipts, invoices, statements, forms and so on. In these instances, it is paramount to detect a large number of relatively small text objects in an image. Further, these objects can be characterized by a large variety of lengths, sizes, and orientations. The challenges faced in this area have been recognized and formalized in  recent ICDAR Robust Reading Competitions: the 2017 challenge focusing on Text Extraction From Biomedical Literature Figures \cite{icdar_cp}, the 2019 challenge focusing on Scanned Receipts OCR and Information Extraction \cite{2019_ICDAR_Challenges_Receipts}, and the 2019 challenge focusing on Arbitrary-Shaped Text \cite{2019_ICDAR_Challenges_ArbShaped} . The images in these challenge sets are characterized by complex arrangements of text bodies scattered throughout an unstructured document and surrounded by "distraction" objects which are not of interest. This regime is one where traditional OCR tools tend to perform poorly and is an area worthy of additional research. 

\begin{figure}
\centering
\includegraphics[width=2.5in]{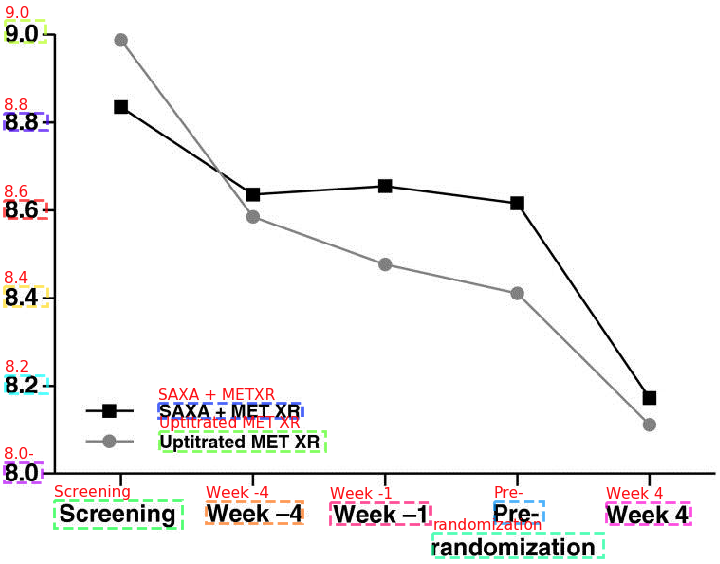}
\caption{Example output on a sample image from ICDAR DeTEXT challenge. Distinct segments of text are identified apart from background pixels and other image objects. The red text on top of each region is the output of the text recognition head.}
\end{figure}

To address this problem, we present a multi-task network that provides an end-to-end trainable model for extraction of text from images of complex documents (see Figure 1). In developing a network for solving these complex OCR tasks, we introduce the following innovations. We describe a multi-task network composed of a convolutional backbone followed by a region proposal network, an object detection head, an object classification head, and a text recognition head.  Additionally, following from work in object detection, we introduce a new pooling mechanism termed RoIRecognitionAlign that is required for instantiating a common representation of image objects that can be processed for text recognition. Additionally, we describe a  convolutional and attentional mechanism for text recognition which we find performs far better than recurrent architectures for this task. Finally, we detail modifications of typical model training procedures that we find are very impactful for establishing the extremely high-accuracy object detection that is required for OCR.

\section{Related Work}

The extraction of text from images can be thought of as a two-step problem: text localization followed by text recognition. In the first part, a model must identify which components of an image correspond to text. The second part then involves "reading" those image segments into sequences of text that are represented within. The problem of text localization shares many features in common with the more general task of object detection, of which we provide a brief overview.

\begin{figure*}
  \includegraphics[width=\textwidth]{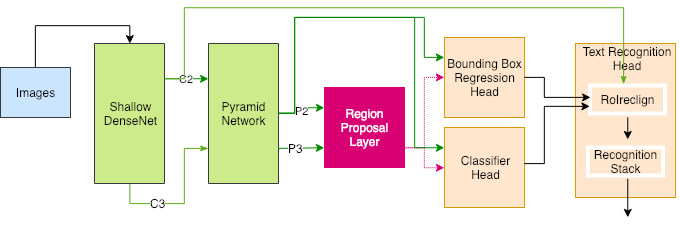}
  \caption{Model Architecture. Image features are extracted through a shared convolutional backbone consisting of a shallow DenseNet and a Feature Pyramid Network. These features are then shared by three model heads: bounding box regression, box classification, and text recognition.  }
  \label{textractor_arch}
\end{figure*}

\subsection{Object Detection}
Recent work in the field of object detection has yielded impressive results in the localization and classification of objects within images. Early work in this area combined Selective Search \cite{Uijlings:2013:SSO:2509349.2509382} with CNN-based methods for image classification \cite{7112511}. To overcome the computational burdens of Selective Search, Regional Convolutional Networks were proposed to leverage convolutional feature map outputs in order to identify Regions of Interests that may contain objects \cite{7112511}. Further improvements over this work focused on computational efficiency \cite{DBLP:journals/corr/RenHG015, DBLP:journals/corr/Girshick15}. For regions proposals, many authors have leveraged the idea of prior boxes (or anchor boxes) and relied on a convolutional regression network to learn the spatial adjustments (change in height, width, centroid) that should be made in order to capture objects \cite{DBLP:journals/corr/Girshick15, DBLP:journals/corr/LiuAESR15, DBLP:journals/corr/RedmonDGF15}.

Recently, Mask R-CNN \cite{DBLP:journals/corr/HeGDG17} combined Region Proposal Networks with a multi-task network that simultaneously solves three problems: bounding box regression, object classification, and identification of a pixel-wise bitmask for image segmentation. This end-to-end network achieves impressive results in object detection and instance segmentation. 

For the task of text localization, we view this as a special case of object detection. Instead of needing to identify dozens or hundreds of distinct kinds of objects, we are primarily only interested in two types of objects: text and background. Therefore, we can utilize the progress made in this general case in order to address challenges in OCR. As we describe in the next section, we can adapt the multi-task learning network described in \cite{DBLP:journals/corr/HeGDG17} to simultaneously solve the problems of text localization and text recognition, thereby benefiting from a shared representation that improves performance in both tasks. 

As was noted by the authors of Mask R-CNN, their model is able to accomplish the multi-task learning challenge of detection and segmentation due largely to their creation of a novel pooling mechanism called RoIAlign. Previous models relied on more lossy pooling mechanisms (RoIPool) \cite{DBLP:journals/corr/RenHG015} which inevitably discards the high-resolution spatial information needed to do pixel-wise instance segmentation. To overcome this, RoIAlign uses interpolation methods to accurately align feature maps with input pixels. As we describe in section 3.4, a different challenge is faced in the domain of text localization and recognition. Here, the recognition component of the model must be able to accurately decode text segments that range in size from one character long to dozens of characters long. If a fixed pooling representation were used to aggregate feature maps (such as in RoIPool), then the high-resolution spatial information needed for accurate decoding would be corrupted. Instead, we develop a new pooling mechanism that is well suited for OCR. 

\subsection{Optical Character Recognition}

In \cite{DBLP:journals/corr/JaderbergSVZ14a}, the problem of text spotting was treated exclusively as a problem of object detection, where the class of objects was as large as the vocabulary of all possible words. In this way, text localization and recognition were combined.  Other authors have used recurrent networks for the text recognition of localized text \cite{DBLP:journals/corr/TianHHH016, Su2014AccurateST, DBLP:journals/corr/HeH0LT15, DBLP:journals/corr/ShiBY15}.  The approach taken by \cite{DBLP:journals/corr/LiaoSBWL16} combines localization and recognition into an end-to-end network. Recent work by \cite{DBLP:journals/corr/abs-1801-01671} also used a multi-task network for text spotting in natural scenes.  Additionally, \cite{DBLP:journals/corr/abs-1803-03474} recently describe a role for an attentional mechanism in text recognition, though one is that different than the approach we describe below. Finally, the challenge of spotting arbitrarily-shaped text has been elegantly approached by \cite{DBLP:journals/corr/abs-1807-02242} and such directions remain as future work for the approach described here.

As described in the next section, our end-to-end multi-task network leverages feature maps that are shared between the localization and recognition branches thus allowing the network to be jointly optimized for solving each task simultaneously. 

\section{Model}
Our network is inspired by Mask R-CNN\cite{DBLP:journals/corr/HeGDG17} and addresses the challenge of OCR with a multi-task network. That is, after the convolutional backbone, feature pyramid and region proposal network, our model has three heads: a localizer (bounding box regression), an text classifier (text or background), and a text recognition network (TRN) head.  For each candidate text-line in an image, our model outputs a bounding box offset (width, height, center), a class label (text or background), and predicted sequence of the text.  A more detailed description of the components follows.

\begin{figure*}
  \includegraphics[width=\textwidth]{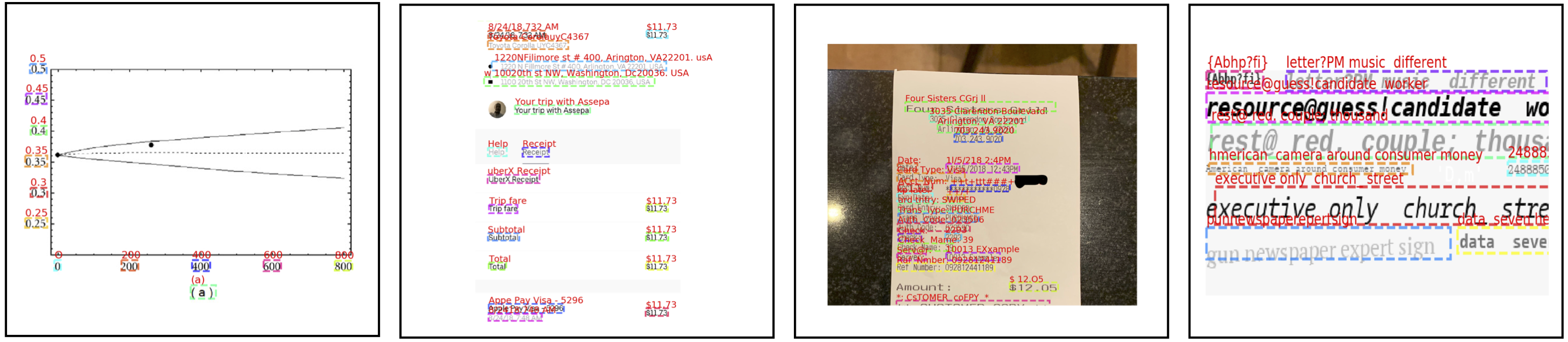}
  \caption{Examples of model output with detected bounding boxes shown as dashed lines and predicted text shown in red. From left to right: (1) an image from the DeTEXT challenge, (2) a screenshot of a receipt from a ride sharing app, (3) a photograph of a receipt, (4) a sample from our synthetic training documents.}
  \label{textractor_examples}
\end{figure*}

\subsection{Backbone}
We replace ResNet\cite{DBLP:journals/corr/HeZRS15} in the backbone, with a customized shallow Densely Connected Network (DenseNet) \cite{DBLP:journals/corr/HuangLW16a} (see Table \ref{tab:architectures}), followed by a Feature Pyramid Network (FPN) \cite{DBLP:journals/corr/LinDGHHB16}. In a convolutional stack such as DenseNet, the features extracted from earlier in the network tend to have high spatial resolution (dense locational details), but have semantically weak features (less relational information about different parts of an image). In contrast, the features extracted from deeper in the network  have lost much of the locational information (low spatial resolution), but they have dense relational information about distant parts of an image (semantically strong features). FPN constructs a top-down architecture that merges the strong features of higher levels of the convolutional stack with the lower ones. In this way, the accuracy of Region Proposal Network will be increased. 
 
\newcommand{\blocka}[4]{\multirow{3}{*}{\(\left[\begin{array}{c}\text{5$\times$5, #1, stride 2}\\\text{3$\times$3, #2}\\ \text{3$\times$3, #3}\end{array}\right]\)conv $\times$1}
}
\newcommand{\blockb}[1]{\multirow{2}{*}{\(\left[\begin{array}{c}\text{1$\times$1}\\ \text{3$\times$3}\end{array}\right]\) conv $\times$ #1}
}
\renewcommand\arraystretch{1.1}
\setlength{\tabcolsep}{3pt}

In evaluating the impact of FPN on our model's overall accuracy, we found that some model components benefited more than others. Output from FPN tends to improve  accuracy in the Region Proposal Network and in the Classifier head. But for the text recognition head, it seems to have a deleterious effect. This is likely due to the demand of very high-accuracy high-resolution information that is required in the case of text recognition. FPN likely adds noise to the basic feature representation, which impairs accuracy in this task. For this reason, the architecture we propose has a blended role for FPN as shown in Figure \ref{textractor_arch}.  The Region Proposal Network, the Classifier and Regression head take input from FPN. More specifically, the RPN receives the whole pyramid (P2 and P3) whereas the Classification and Regression heads receive just P2. In contrast, the text recognition head gets its input directly from the convolutional backbone (level C2 as described below).

\subsection{Text Localization}\label{text_localization}

A primary difference between the challenges posed in object detection and challenges posed in OCR comes from the large variety of aspect ratios inherent to blocks of text. The height of a text block is determined by its font size, but the width depends on both its font size and the length of the text content being represented (number of characters). Due to this, it is not uncommon for a single image to contain text blocks composed of only a few characters as well as text blocks containing long phrases and sentences. Being able to accommodate this diversity of object sizes and aspect ratios (all within the same object class)  is an important divergence from typical tasks in object detection. We made changes to both the anchor boxes' aspect ratios and the pooling dimensions of RoIAlign. We set the anchor boxes' aspect ratios to \textit{(1, 2, 4)}, which is suitable for the challenge datasets we use for evaluation (see Results). We also implement oriented bounding box detection, similar to recent approaches \cite{DBLP:journals/corr/abs-1801-01671, DBLP:journals/corr/abs-1711-09405}. Here, the region proposal network will predict not just spatial coordinates for a box's center, width, and height $(c_x, c_y, w,h)$, but additionally angle of rotation $\theta$. This angle is then used to rotate the RoI with a linear projection, for which the corresponding features are sampled using bilinear interpolation \cite{DBLP:journals/corr/abs-1803-03474}.

\subsection{Text Recognition Network (TRN)}
One of the main contributions of this paper is the added text recognition head, which plays a similar role as the masking head in Mask R-CNN. The TRN head consists of two main components: a new pooling mechanism tailored for text recognition (RoIRecognitionAlign), and an OCR stack. The following  subsections detail TRN's components. 

\subsubsection{RoI Recognition Align}
\label{roialign} As mentioned in section \ref{text_localization}, in OCR we only have two classes: text, and background. As a result we are assigning many independent lines of texts to the same class, no matter how long they are, or what their font sizes are. Due to this, we encounter a large variety of aspect ratios for the same class of object, and we cannot treat them the same way.

In Mask R-CNN, large objects are separated from small ones by pooling features of larger objects from the top of the feature pyramid, and pooling features of small objects from the bottom of the pyramid. Since top levels of the feature pyramid have greater strides, it is sufficient to re-size extracted features for all the objects to the same dimensions. However, to distinguish different characters in an image, high spatial resolution  details need to be available in the feature maps. Thus, decoding text places a more stringent requirement on feature representation than is required for object detection alone. As a result, the previously used techniques of pooling the features of larger object (texts) from the top of feature pyramid cannot be applied here. This would result in low spatial resolution information and would degrade the models ability to accurately recognize the characters. 

Considering these challenges, we designed RoIreclign to pool the features of every RoI, no matter how big or small, from the second convolutional block of ResNet (C2). This was chosen to retain the high spatial resolution information that would be required to accurately recognize the characters of localized texts. We have found that C2 has the best balance of abstract and specially detailed features, required for the text recognition task.

Further, we must address the problem of diverse text aspect ratios as we do not want our pooled representation to be corrupting text features  by stretching short texts and compressing long ones. Our solution is to dynamically re-size and pad all RoIs into a fixed shape. Specifically, we define a fix height ($H_{o}$) and width($W_{o}$) for the output of RoIreclign. Based on the output height and the height of each RoI ($H_{roi}$), we calculate its re-sized width ($W_{r}$), and re-size the extracted features to the output height and the calculated width. 
\begin{equation}\label{eqn1}
	H_{r} = H_{o}
\end{equation}
\begin{equation}\label{eqn2}
    W_{r} = W_{roi} \times \frac{H_{r}}{H_{roi}}
\end{equation}
\begin{equation}\label{eqn3}
    W_{p} = W_{o} - W_{r} 
\end{equation}
Padding width ($W_{p}$) of each RoI is calculated based the output width, and the re-sized features will be padded if required. The output width dictates the maximum number of characters that the model can detect for any given line of text. In this way, we preserve the natural aspect ratio of the extracted features and do not warp them. In implementation, our typical pooling height and width is either ($5\times180$), ($6\times224$) or (7, 300).

By having the dynamic re-sizing, in a way we are re-scaling every text to a consistent representational font size, as a result the same architecture can be used to recognize characters of every localized line. Dynamic re-sizing is accomplishing a similar goal as extracting features from different levels of feature pyramid based on the size of the RoI. 

\begin{figure}
  \includegraphics[width=1.0\linewidth]{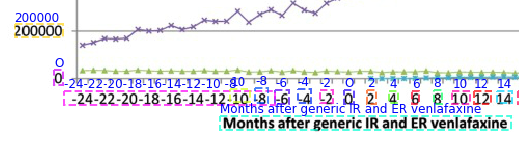}
  \caption{Example of model output where it failed to distinguish independent values. There are two main issues here: inconsistent white-space, and very small white-space between text entities.}
  \label{textractor_failure_examples}
\end{figure}

\subsubsection{Recognition Stack}

A common approach to text recognition within each bounding box is cropping the input image according to predicted bounding boxes and using that image crop as the input to the OCR network. In contrast, our end to end solution uses generated features from the convolutional backbone as TRN's input.  This provides a richer feature set for TRN and, as a result, it is possible to accomplish the recognition task with a shallower network. More importantly, the OCR's accuracy will be propagated to the backbone and adjust the features accordingly. This ensures that the representations leveraged by all three heads take into account the tasks at each of the heads, allowing the heads to jointly optimize each other and not work independently.  As mentioned in section \ref{roialign}, in order to recognize the text, we are using the second convolutional block of the convolutional backbone, and pull the predicted RoIs from these feature maps. 

The feature maps from the convolutional backbone are initially transformed through a convolution that is specific to TRN: the RoIreclign layer described previously. For each RoI, this results in a consistently-shaped feature map of shape (\emph{Height} x \emph{Width} x \emph{Channels}), which in our implementation is either ($5\times 180\times 256$), ($6\times 200\times 256$) or ($7\times 300\times 256$). In all cases, the Width represented in this pooling layer is tuned to a high upper-bound on the number of characters typically found in an RoI. This representation for each RoI is then input into the recognition module, the output of which is a predicted sequence over the model's alphabet. The model's alphabet contains $A$ distinct tokens comprising letters, numbers, and a variety of punctuation and special characters. 

Instead of an RNN architecture for text recognition, we have found improved performance using a  convolutional stack and we need further improvement utilizing a supervised attention mechanism described below. Additional details about different architectures and their performances are provided in table \ref{tab:architectures}. \footnote{The results Table \ref{tab:architectures} are computed with a evaluation metric that closely resembles \cite{COCO_eval_page}, whereas the evaluation metric reported in Table \ref{tab:evaluation} are calculated from the DeTEXT submission site on the testing set \cite{icdar_cp}. Therefore, Table \ref{tab:architectures} and Table \ref{tab:evaluation} should not be directly compared.}  Overall, we speculate that for text recognition, it is only local spatial dependencies that are important for accurately transcribing each character. Therefore, a convolutional method with small-range kernels can be expected to be perform adequately as compared to an RNN approach which considers the entire spatial extent of an RoI.

\setlength{\tabcolsep}{3pt}
\begin{table*}[t]
\begin{center}
\begin{tabular}{ll|c|c|c|c|}
\cline{3-6}
 &  & \multicolumn{4}{c|}{\textbf{Dataset}} \\ \cline{3-6} 
 &  & \multicolumn{2}{c|}{\textbf{ICDAR Validation Set}} & \multicolumn{2}{c|}{\textbf{Synthetic}} \\ \hline
\multicolumn{1}{|l|}{\textbf{Architecture}} & \textbf{Variation} & \multicolumn{1}{l|}{\textbf{BBox mAP}} & \multicolumn{1}{l|}{\textbf{f-score}} & \textbf{BBox mAP} & \textbf{f-score} \\ \hline
\multicolumn{1}{|l|}{RNN} & \multicolumn{1}{c|}{\multirow{5}{*}{\begin{tabular}[c]{@{}c@{}}TRN\\ Head\end{tabular}}} & 0.57 & 11.8 & 0.72 & 26.4 \\ \cline{1-1} \cline{3-6} 
\multicolumn{1}{|l|}{CONV} & \multicolumn{1}{c|}{} & 0.64 & 11.1 & 0.95 & 30.5 \\ \cline{1-1} \cline{3-6} 
\multicolumn{1}{|l|}{Unsupervised Attn-Concat} & \multicolumn{1}{c|}{} & 0.62 & 17.6 & 0.83 & 37.2 \\ \cline{1-1} \cline{3-6} 
\multicolumn{1}{|l|}{Unsupervised Attn-Multi} & \multicolumn{1}{c|}{} & 0.60 & 19.8 & 0.80 & 36.4 \\ \hline \hline
\multicolumn{1}{|l|}{Custom Shallow ResNet} & \multirow{2}{*}{Backbone} & 0.70 & 16.8 & 0.95 & 41.2 \\ \cline{1-1} \cline{3-6} 
\multicolumn{1}{|l|}{Custom Shallow DenseNet} &  & 0.67 & 23.1 & 0.95 & 52.3 \\ \hline
\end{tabular}
\end{center}
\vspace{-.5em}
\caption{Comparison of architectures for recognition head (top) and for convolutional backbones (bottom). In \textbf{Unsupervised Attn-Concat} the sigmoid vector is concatenated to the end of the feature map, whereas in \textbf{Unsupervised Attn-Multi} the sigmoid vector (replicated per each channel) is multiplied by feature map.
(replicated per each channel)}
\label{tab:architectures}
\vspace{-.5em}
\end{table*}

\textbf{\emph{Convolution}}
For each RoI, a ($H\times W\times C$) feature map is output from RoIreclign. We then convolve this feature map with kernels of height H and of small-range width to produce a map of shape ($1\times W\times C$) representing C features at each of W potential character locations in the RoI. At each spatial location, a probability vector over the alphabet is computed using a shared Dense layer of size ($C\times A$). A CTC loss is used to compress the length-W output sequence to a reasonable predicted target sequence.

\textbf{\emph{Attention}}
In additional to the convolutional mechanism described above, we also introduce an attentional mechanism that helps the recognition head focus on spatially-relevant features for predicting each character. The input to the attention module is the ($1\times W\times C$) feature map and the output is a ($1\times W$) attention vector, $\vec{a}$. 

The attention vector is used to re-weight the ($1\times W\times C$) feature map.  A pointwise product is use to scale spatial location of the feature map according to the attentional weights. This allows the model to de-emphasize regions of probably whitespace and to focus on regions containing characters.

\begin{figure}
\centering
  \includegraphics[width=0.6\linewidth]{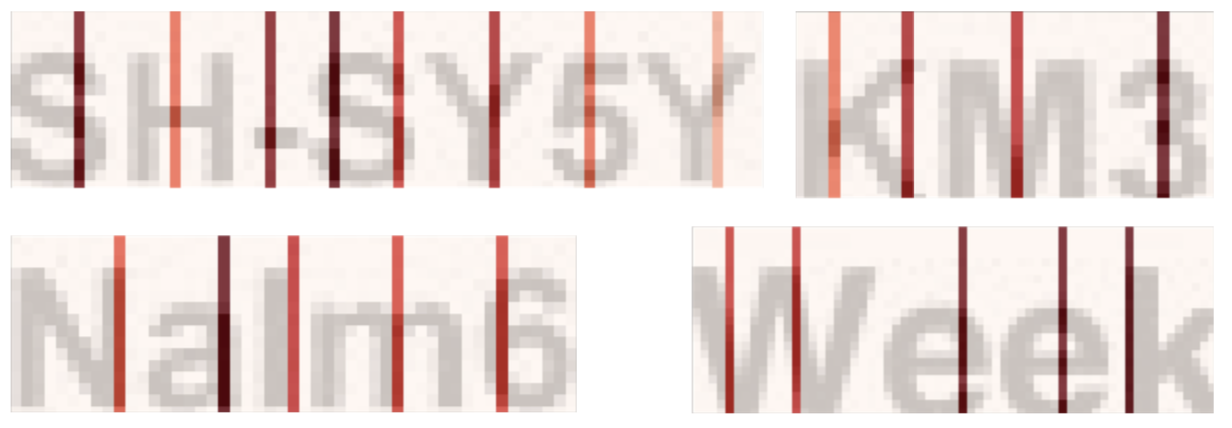}
  \caption{Visualization of attention vectors at inference time for several RoIs. The peak of the attention vectors are visualized in red and overlaid on the input image space. Note that the model attends to the central portion of most characters, or to multiple locations for wide characters. Additionally, the number of attentional peaks tends to mirror the number of characters in an RoI.}
  \label{attn_fig}
\end{figure}

In order to train this attention mechanism, two approaches were explored. The first, which we refer to as "unsupervised attention" allows the free parameters of the attentional mechanisms to be learned in an unconstrained way, with no additional components of the recognition loss. This approach is adequate and results are described in Table \ref{tab:architectures}. However, to more efficiently train the attention mechanism, we can also provide an additional labeled datasource, which considerably improves the performance. At training time, the attention module receives additional input which indicates the true locations of characters within each ground truth box. Specifically, a ($1\times W$) binary mask is generated for each text box. This mask indicates which of the $W$ spatial locations corresponds to the center of any character within the box. A sigmoid loss is used to compare the predicted attention vector $\vec{a}$ with the ground-truth mask vector. This training mechanism improves overall recognition performance and is reflected in results reported in table \ref{tab:evaluation} as well as Figures 1 through 5.

Internally, the attention module can be computed in many ways such that a ($1\times W$) attention vector is produced. In our implementation we use a short-range convolution  followed by a shared Dense layer with sigmoid output. This allows the attention module to consider local spatial correlations while computing whether each spatial location is likely relevant to character recognition or whether it corresponds to whitespace. Figure \ref{attn_fig} shows visualizations of attention vectors at inference time for several RoIs. The attentional peaks are shown in red and it can be seen that they send to align with the centroid of typical characters and that the number of such peaks will mirror the number of characters in an RoI.

\subsection{Loss}
We follow typical loss functions as seen in previous multi-task object detection work \cite{DBLP:journals/corr/HeGDG17}. The primary difference comes from the inclusion of text recognition head, for which we use a CTC loss between the predicted text sequence and the ground-truth sequence \cite{Graves:2006:CTC:1143844.1143891}. We also have the attentional loss described above which focuses the Recognition head on the centers of characters within RoIs. The total loss is the sum of the individual losses.

\subsection{Synthetic Training Data}

Training our end-to-end model requires the creation of  synthetically-generated training documents for which we have full ground truth. Relying on hand-labeled documents would prove too time consuming to reach a large enough sample size. Therefore, we chose to generate synthetic documents that retain many of the features inherent to real world complex documents. This includes using more than 40 fonts, many font sizes, randomized text segments, lengths, and mathematical symbols, randomized text colors and background colors, sizes and gaps between text elements, randomized text orientation and tilt, image noise and blur, as well as the inclusion of "distraction" objects such as geometric shapes, lines and graphs.

\subsection{Training Procedure}

The training procedure for the models described here consisted of 10,000 synthetic document images with a batch size of 2. Model parameters are estimated with SGD with momentum: a learning rate of 0.001 with momentum of 0.9 was used for the first 700 epochs and was then  switched to momentum of 0.5. 

For evaluation with the ICDAR dataset (see below), the model was then fine-tuned with the provided training dataset (100 images) for 10 additional epochs.

\section{Results}
We validate our approach using the 2017 ICDAR Robust Reading Competition (\url{http://rrc.cvc.uab.es/}) DeTEXT - Text Extraction From Bio-medical Literature Figures \cite{DBLP:journals/plos/Yin2015}. In this work, we are deliberately interested in recognizing text from images of complex documents. For these reasons, a fitting validation task for our approach is the Biomedical Literature Figures challenge. Here, the corpus of images contain scientific figures, from which we are to extract any text from annotations, legends, or axes labels.

In addition to the ICDAR challenge data-sets, we also evaluate our methods against synthetically generated documents for which we have complete ground truth of text locations and transcripts. These documents are meant to mimic many real-word challenges faced in complex documents such as receipts, invoices, statements, and so on. Further detail about the nature of these synthetic documents is described in section 3.6.  For evaluation, new batches of unseen synthetic documents are generated. 

To our knowledge, there are no published studies which report results on the DeText Biomedical Literature Figures challenge. For this reason, it is difficult to make direct comparisons with existing methods. While there exist submissions to the public competition page for the challenge, we are unable to evaluate or understand the methods underlying those submissions, as they are not accompanied by publications. As described below, we can benchmark our performance relative to other submissions, even where ours are not top-ranked. However, we cannot speak to the true implementation of other methods, which seem to be based on Object Detection methods (similar to what is described here) but also occasionally indicate the presence of post-processing methods, the details of which are unknown to us. 

Finally, we mention that our focus so far has been in the establishment of a multi-task learning framework for these challenges in non-traditional OCR, but many future enhancement remain. In particular, the DeText challenge images are characterized by many rotated and off-horizontal text elements. It is certainly an important area of future work to incorporate rotational bounding boxes or similar \cite{DBLP:journals/corr/abs-1801-01671, DBLP:journals/corr/abs-1807-02242}.  Therefore, our  results reported here are limited by an initial focus on non-oriented text, and this limitation leads to inaccuracies for rotated text boxes. 

\subsection{Evaluation Metrics}
Evaluation of model performance is conducted through the ICDAR DeText competition page \cite{icdar_cp} evaluation application. This application provides the ability to evaluate performance on each of the challenge tasks including Localization, Cropped Text Recognition, and End-To-End Recognition.  Evaluation metrics are similar to those defined in the COCO-Text challenge \cite{COCO_eval_page}, briefly described below.

\emph{Localization}
For Localization in particular, a detection is considered a True Positive if the detected bounding box overlaps with the ground truth bounding box with a Intersection-over-Union (IOU) of pixel overlap which exceeds a given threshold. Aggregated over all RoIs and all images, the net result is reported as mAP for Localization. 

\emph{End-To-End Recognition}
For Recognition, a True Positive is attained if (1) the detected bounding box has sufficient overlap with the ground truth bounding box and (2) the predicted output text matches perfectly the ground truth label text. This requirement for perfect Recognition is especially challenging in the absence of a lexicon, for which there is none in the Biomedical Literature Figures challenge. 

\begin{table}[]
\begin{tabular}{l|c|c|}
\cline{2-3}
 & \multicolumn{2}{c|}{\textbf{ICDAR 2017 DeText on Testing Set}} \\ \hline
\multicolumn{1}{|c|}{\multirow{2}{*}{\textbf{Model Type}}} & \multirow{2}{*}{\textbf{\begin{tabular}[c]{@{}c@{}}Localization\\ (mAP)\end{tabular}}} & \multirow{2}{*}{\textbf{\begin{tabular}[c]{@{}c@{}}End-to-End Recognition \\ (mAP)\end{tabular}}} \\
\multicolumn{1}{|c|}{} &  &  \\ \hline
\multicolumn{1}{|l|}{Non-Rotational} & 35.5 & 9.5 \\ \hline
\multicolumn{1}{|l|}{Non-Rotational with Attention} & 43.3 & 17.2 \\ \hline
\multicolumn{1}{|l|}{Rotational with Attention} & 36.9 & 5.8 \\ \hline
\end{tabular}
\vspace{-.5em}
\caption{Model performance on 2017 DeText Biomedical Literature Figures challenge test set. }
\vspace{-.5em}
\label{tab:evaluation}
\end{table}

\subsection{Comparison of Architectures}
Table \ref{tab:evaluation} provides a comparison of modeling approaches across the DeText benchmark, for the end-to-end model with and without the attentional mechanism. The  model shown here is the DenseNet backbone with a Convolutional TRN, trained for 1000 epochs with synthetic documents. The model was then fine-tuned with an additional 10 epochs of training on the ICDAR DeTEXT training dataset (100 images). We note that even modest fine-tuning with the ICDAR training set leads to considerable performance improvement, relative to training with synthetic documents alone.

As noted previously, comparison of our results with previous researchers remains challenging. On the DeText competition page, there exist a small number of submissions for the Localization and End-to-End Recognition tasks. When compared to these, our model is outperformed by the leaders in each of these, where we see Localization $mAP$ in the range of 0.9 at the best and End-To-End Recognition in the range of 0.6 at the best. However, we are unable to comment on the primary differences here, as these submissions are not accompanied by publications. While surely these methods adopt similar advances in Object Detection and multi-task learning as we do here, it is unclear what post-processing and other enhancements might be at the heart of these methods. To our knowledge, the evaluations presented here are the first published results on the DeText Biomedical Literature Figures challenge.

As shown in Table \ref{tab:evaluation}, our primary modes for evaluation are comparing three main architectural choices. The "Non-Rotational" model is the model as described in section 3.2, except without the inclusion of rotation estimation in the region proposal network. Therefore this model is expected to be most performant with horizontal text boxes, or perhaps boxes with very slight off-horizontal orientation. The "Non-Rotational with Attention" model extends the previously described model with the attentional mechanism described in 3.2.2, which aims to improve recognition performance. The most general model, "Rotational with Attention", allows the model to accommodate arbitrarily oriented text and also includes the attentional mechanism in the recognition head. 
 
In both Localization and Recognition, it is clear that the attention mechanism provides an important enhancement. Not surprisingly, this is most impactful for Recognition, where the $mAP$ is approximately doubled with the addition of attention.  In contrast, the flexible model with rotational bounding boxes, which will be required for generalized text spotting, yields a diminished efficacy in comparison to a non-rotational approach. While the reason for this remains unclear, we note that requirement for oriented text yields new challenges for model training and the learned representations.

\section{Conclusion}
While previous work has achieved impressive results in traditional OCR and in scene-based text detection, regimes of non-traditional OCR present outstanding challenges for the field. In this regime, our models must be able to accommodate arbitrarily complicated layouts of text, diverse text size, shapes, and rotations. This particular challenge was well-formalized in the 2017 ICDAR DeTEXT Biomedical Literature Figures challenge. Real-world use cases include OCR of complex documents such as receipts, forms, statements, and so on. We described a multi-task network that combines object detection with text recognition in order to extract text from complex documents without a lexicon. We presented a novel pooling mechanism (RoIreclign) that is required to retain high-spatial resolution information that is needed for accurate text recognition. We evaluated several architectures for text recognition and find that convolutional and attentional mechanisms strongly out-perform the more common recurrent approaches.  Our approach is flexible and is easily optimized to particular use cases with a modest amount of fine-tuning.

\section*{Acknowledgments}

The authors would like to thank anonymous reviewers for helpful feedback on this work, as well as colleagues in Capital One's Center for Machine Learning for thoughtful discussions.

\bibliography{bibliographies/textractor}{}
\bibliographystyle{plain}

\end{document}